\documentclass[11pt, a4paper]{article}

\usepackage[utf8]{inputenc}
\usepackage{amsmath}
\usepackage{amssymb}
\usepackage{graphicx}
\usepackage[margin=1in]{geometry}
\usepackage{hyperref}
\usepackage{algorithm}
\usepackage{algpseudocode}
\usepackage{booktabs}
\usepackage{tabularx}
\usepackage{longtable}
\usepackage{algorithm}

\usepackage{multirow}   
\usepackage{subcaption}
\usepackage[table]{xcolor} 
\usepackage{booktabs} 

\usepackage{geometry}
\hypersetup{
    colorlinks=true,
    linkcolor=blue,
    filecolor=magenta,      
    urlcolor=cyan,
    pdftitle={Prompt-Response Semantic Mutual Information for Hallucination Detection},
    pdfpagemode=FullScreen,
}

\title{\textbf{Prompt-Response Semantic Divergence Metrics  for Faithfulness Hallucination and Misalignment Detection \\
in Large Language Models}}

\author{Igor Halperin\thanks{The author acknowledges the assistance of Gemini 2.5 Pro, Claude 4 Opus and Grok 3 in the preparation of this manuscript and its associated code. 
Some of content hallucinations created by AI assistants were caught and removed by the author.
All remaining errors are the author's own. The views expressed herein are those of the author and do not necessarily reflect the views of his employer. Email for correspondence: ighalp@gmail.com.} \\ Fidelity Investments}
\date{\today}

\begin{document}

\maketitle

\begin{abstract}
The proliferation of Large Language Models (LLMs) is challenged by hallucinations, critical failure modes where models generate non-factual, nonsensical or unfaithful text. This paper introduces Semantic Divergence Metrics (SDM), a novel lightweight framework for detecting Faithfulness Hallucinations - events of severe deviations of LLMs responses from input contexts. We focus on a specific implementation of these LLM errors, \textbf{confabulations}, defined as responses that are  arbitrary and semantically misaligned with the user's query.

Existing methods like Semantic Entropy test for arbitrariness by measuring the diversity of answers to a single, fixed prompt. Our SDM framework improves upon this by being more prompt-aware: we test for a deeper form of arbitrariness by measuring response consistency not only across multiple answers but also across multiple, semantically-equivalent \textit{paraphrases} of the original prompt. Methodologically, our approach uses joint clustering on sentence embeddings to create a shared topic space for prompts and answers. 
A heatmap of topic co-occurances between prompts and responses can be viewed as a quantified two-dimensional visualization of the user-machine dialogue.
We then compute a suite of information-theoretic metrics to measure the semantic divergence between prompts and responses. Our practical score, $\mathcal{S}_H$, combines the Jensen-Shannon divergence and Wasserstein distance to quantify this divergence, with a high score indicating a Faithfulness hallucination. Furthermore, we identify the KL divergence KL(Answer $||$ Prompt) as a powerful indicator of \textbf{Semantic Exploration}, a key signal for distinguishing different generative behaviors. These metrics are further combined into the {\bf Semantic Box}, a diagnostic framework for classifying LLM response types, including the dangerous, confident confabulation. Our work provides a principled, prompt-aware methodology for real-time detecting faithfulness hallucinations 
and semantic misalignment in black-box LLMs, for either a single query or a multi-turn conversation.

\end{abstract}

\section{Introduction}

Large Language Models (LLMs) like Gemini, Claude, and GPT-4 have demonstrated remarkable capabilities in generating human-like text. However, their reliability is undermined by critical failure modes known as LLM hallucinations \cite{zhang2023siren}. A comprehensive taxonomy of these errors distinguishes between \textit{intrinsic} hallucinations (contradicting the provided 
input or context) and \textit{extrinsic} hallucinations (claims that cannot be verified against the source). Another key axis distinguishes between \textit{Factuality} (alignment with real-world knowledge) and \textit{Faithfulness} (alignment with the user's prompt and provided content) \cite{cossio2025taxonomy}.

In this paper, we are primarily concerned with detecting \textbf{intrinsic faithfulness hallucinations}. We frame this as a problem of measuring the \textbf{semantic divergence} between an LLM's response and the full input context provided by the user.

Some researchers have proposed that the term \textbf{confabulation} is more appropriate than "hallucination" for describing these LLM failures \cite{farquhar2024semantic}. In psychiatry, "hallucinations" typically refer to sensory-related phenomena, while "confabulations" are false or distorted memories that a person believes to be true. This occurs when an individual fills in gaps in their memory with fabricated details without the conscious intent to deceive. This psychiatric definition makes "confabulation" a meaningful synonym for the kind of contextual misalignment or unfaithfulness of LLM responses that we aim to detect.

For LLM applications, one way to define confabulations is to say that it means LLMs outputs that are both incorrect and arbitrary, while their arbitrariness is due to their excessive sensitivity to the statistical noise in LLM's output generation. 
The Semantic Entropy (SE) method \cite{farquhar2024semantic} builds on this idea, and defines LLM confabulations as the sensitivity of the LLM's output to inessential details like the random seed used for temperature sampling.
The SE method amounts to multiple sampling of LLM responses with different random seed while keeping the same initial prompt for each LLM call.
Variability of produced answers is estimated by performing semantic clustering and measuring entropy of the resulting clustering distribution for all LLM responses. This results in their final score, Semantic Entropy (SE). 


While powerful, the SE method has a key limitation: it is not (sufficiently) prompt-aware. The information about the prompt is not fully utilized, while the context is simply used to anchor each LLM response but not in any other way. 
The prompt remains a fixed contextual background.\footnote{Because the SE method measures response diversity without deeply conditioning on the prompt's content, it functions primarily as a detector for a type of \textit{extrinsic} hallucination.}
The SE method only tests for instability with respect to repetitions of the fixed prompt without accessing its context, and can mistakenly flag a legitimately complex answer actually implied by the prompt as a confabulation.

In this paper, we use a wider interpretation of LLM confabulations which makes them more similar to  intrinsic faithfulness hallucinations.\footnote{For this reason, we will use the terms 'confabulations' and 'intrinsic faithfulness hallucinations' interchangeably.} 
We define confabulations as  
incorrectness and arbitrariness of LLM responses \textbf{relatively to the provided input context}. A confabulation, in our framework, is a response that is semantically unrelated, completely or partially, to the user's prompt---whether that prompt is a simple query or a query augmented with extensive reference documents.

Our central thesis is that a faithful, non-confabulatory response should remain semantically aligned with the world defined by its prompt. A confabulation, therefore, can be understood as a severe form of {\bf semantic misalignment}. We propose a novel lightweight framework, \textbf{Semantic Divergence Metrics (SDM)}, to directly quantify this alignment in a more robust, prompt-aware manner. The framework is designed for the analysis of dynamic prompt-answer pairs and is applicable to both single interactions and multi-turn conversations with an LLM.

The SDM methodology operates in a black-box setting, and amounts to a two-step procedure. First, it probes for arbitrariness more deeply by generating multiple responses to a set of semantically equivalent \textbf{paraphrased prompts}. Second, our approach uses \textbf{joint embedding clustering} to establish a shared topic space for prompts and answers. From there, we compute a suite of information-theoretic and distributional metrics. Our first final score, $\mathcal{S}_H$, combines the Jensen-Shannon divergence (JSD) and Wasserstein distance to provide a robust, single-value measure of semantic drift, where a high score indicates a higher risk of confabulation. Furthermore, we identify the KL divergence KL(Answer $||$ Prompt) as a powerful measure of \textbf{Semantic Exploration}, a key signal for distinguishing between different generative behaviors. This metric constitutes the second final score of our SDM method.

Through a series of controlled experiments, we show that our framework successfully distinguishes between different types of responses. We use a visual heatmap analysis (Section~\ref{sec:experiments_viz}) to reveal the distinct visual signatures of different alignment strategies. Combining our metrics, we introduce the \textbf{Semantic Box} (Section~\ref{sec:semantic_box}), a novel diagnostic framework that uses our metrics to classify response behaviors—from faithful alignment to the most dangerous, confidently misaligned falsehoods. This work presents a principled, quantitative approach to measuring a critical aspect of LLM reliability: their ability to produce responses that are semantically aligned with a user's query.

Our contributions are threefold:
\begin{enumerate}
    \item We introduce a novel, prompt-aware method for confabulation detection that tests for arbitrariness more robustly using paraphrased prompts.
    \item We propose a simple practical algorithm for measuring semantic alignment based on joint clustering of prompt-answer embeddings and the resulting divergence metrics (Ensemble JSD, KL, etc.).
    \item We develop a comprehensive diagnostic framework, the Semantic Box, which uses these metrics to classify different modes of LLM behavior, including the confident confabulation, a critical failure mode.
\end{enumerate}

The remainder of this paper is structured as follows. Section 2 reviews related work in hallucination detection. Section 3 details the foundational Semantic Entropy method. Section 4 presents our proposed SDM methodology. Section 5 analyzes its computational complexity. Section 6 presents our experimental results. Section 7 introduces our visualization tool, the Semantic Box. Section 8 presents a discussion of our results. Section 9 outlines future work  and research directions, and Section 10 concludes.

\section{Related Work}

Hallucination detection in LLMs is a rapidly evolving field. Research can be broadly categorized by the information source used for verification. Intrinsic methods check for contradictions within the generated text itself, while extrinsic methods validate claims against an external knowledge base \cite{ji2023survey}. Our work focuses on detecting a specific type of faithfulness hallucination---the confabulation---in black-box settings, where only the model's inputs and outputs are accessible.

A primary approach in black-box settings is to measure \textbf{response consistency}. Methods like SelfCheckGPT \cite{manakul2023selfcheckgpt} sample multiple responses to a single prompt and check for consistency. The concept of Semantic Entropy \cite{farquhar2024semantic} refines this by measuring the semantic diversity of the answer set, treating high entropy as a signal of arbitrariness, a key component of confabulation. These methods, however, are prompt-agnostic; they only test for instability with respect to a single, fixed phrasing of a question.

A more sophisticated approach involves testing the model's robustness against transformations of the input prompt. The framework of \textbf{metamorphic testing} has been successfully applied to this problem. In their work on \textbf{MetaQA}, Yang et al. \cite{yang2024hallucination} introduce a set of metamorphic relations---transformations applied to a prompt that should logically lead to a predictable transformation in the answer. For example, paraphrasing a prompt (a "meaning-preserving" transformation) should ideally yield a semantically equivalent answer. By comparing the original answer to the answer from the transformed prompt, they can detect inconsistencies that signal a potential hallucination.

Our SDM framework shares the core insight of using prompt transformations, specifically paraphrasing, to probe for instability. However, our methodology differs in two fundamental ways. First, where MetaQA typically compares a single original answer to a single transformed answer, our approach uses an \textbf{ensemble-based analysis}. We generate multiple responses for \textit{each} of the paraphrased prompts, allowing us to measure the statistical properties of the entire response distribution, not just a single instance. Second, our method of comparison of prompt-response pairs is different. Rather than relying on the LLM-as-a-judge method, we use \textbf{joint embedding clustering} to create a shared, sentence-level topic space. This allows us to employ a richer suite of information-theoretic and distributional metrics (JSD, KL Divergence, Wasserstein Distance) to provide a more granular and multi-faceted measure of semantic divergence and the risk of confabulation.

Other works have also used embeddings to measure prompt-response alignment. A recent approach, \textbf{SINdex} \cite{abdaljalil2025sindex}, proposes concatenating the input prompt with a generated response into a single string and then computing a single embedding for this combined text. By comparing the embeddings of multiple such `[prompt + response]` strings, it measures inconsistency. This operates at the document level, treating the entire text as a single semantic unit. In contrast, our SDM method performs a more granular, \textbf{sentence-level analysis}. By jointly clustering the individual sentences of both prompts and answers, we construct a shared topic space that allows for a much finer-grained measurement of topic divergence and content drift, as captured by our JSD and Wasserstein metrics. Ref.~\cite{agrawal2024language} studies how prompt complexity affects model behavior but do not directly address hallucination detection.

\section{Background: The Semantic Entropy Method}

The Semantic Entropy (SE) approach, introduced by Farquhar et al. \cite{farquhar2024semantic}, provides the conceptual foundation for our work. It aims to detect confabulations by measuring the "arbitrariness," or semantic inconsistency, within a set of LLM-generated answers to a single prompt. The key steps of their method are:

\begin{enumerate}
    \item \textbf{Generate Multiple Answers:} For a given prompt, generate $N$ different answers using temperature-based sampling.
    
    \item \textbf{Sentence Segmentation:} Each answer is broken down into its constituent sentences.
    
    \item \textbf{NLI-Based Clustering:} The core of the SE method is its clustering process. It clusters sentences based on semantic equivalence. A Natural Language Inference (NLI) model is used to determine if pairs of sentences bidirectionally entail each other. Sentences are grouped into the same cluster if they are determined to be semantically equivalent by the NLI model. Each resulting cluster represents a distinct "topic" (or "semantic concept").
    
    \item \textbf{Entropy Calculation:} The cluster assignments form a discrete probability distribution, where $p_i$ is the fraction of sentences belonging to cluster $i$. The Shannon entropy is then calculated as $H = -\sum_{i=1}^{k} p_i \log_2 p_i$.
    
    \item \textbf{Hallucination Detection:} A high SE score suggests that the model is generating many inconsistent or contradictory ideas for the same prompt, signaling a higher likelihood of confabulation.
\end{enumerate}

While this is a powerful method for measuring response diversity, its primary drawback is its largely prompt-agnostic nature. It cannot distinguish between the high entropy of a confabulated response and the legitimate high entropy of a correct, multi-faceted answer to a complex prompt. Our work builds on this idea of measuring response diversity but introduces prompt-awareness and uses a different technological approach (joint embedding clustering) to achieve a more nuanced, context-aware analysis.

\section{Methodology: Semantic Divergence Metrics (SDM)}

Our proposed SDM framework is designed to detect a specific, challenging type of hallucination: the \textbf{confabulation} which is defined  as a LLM response that makes claims that are both incorrect and arbitrary \textbf{relative to the provided input context}. This input context, which we will refer to generally as the "prompt," defines the complete ground-truth semantic space for a faithful response. 
Our framework's core principle is that a well-aligned, non-confabulatory response must remain semantically consistent with this prompt-defined world. This section details our approach, starting with the information-theoretic foundation that motivates our method and culminating in designing practically-oriented signals for assessing the likelihood of confabulation in a given prompt-response pair.

\subsection{Information-Theoretic Foundation: A Diagnostic Framework for Confabulations}

The foundation of our SDM framework is the principle that the complete input provided to an LLM---the prompt---defines the ground-truth semantic space for a well-aligned, non-confabulatory response. A confabulation is therefore a deviation from this specific, prompt-defined semantic space. Our framework is designed to be general, treating the "prompt" as the entirety of the context provided to the model. This prompt context can take two primary forms:

\begin{itemize}
    \item \textbf{A direct query,} where the prompt itself is the complete context (e.g., "Compare and contrast the economic policies of Keynes and Hayek.").
    \item \textbf{A query augmented with reference documents,} where the prompt includes external information that the model is expected to use (e.g., in Retrieval-Augmented Generation, RAG).
\end{itemize}

Instead of relying on a single metric, as e.g. is done in the Semantic Entropy method, we propose that confabulations can be diagnosed along two orthogonal dimensions, analogous to the concepts of \textbf{bias and variance} in machine learning.

\begin{enumerate}
    \item \textbf{Semantic Exploration (The "Bias" Dimension):} The first dimension measures the semantic drift or "bias" of the answer distribution away from the prompt distribution. It quantifies the degree to which the LLM must elaborate, invent, or extrapolate beyond the concepts explicitly provided in the prompt. We measure this using the Kullback-Leibler (KL) divergence KL(Answer $||$ Prompt). A high value for this \textbf{Semantic Exploration} metric indicates that the model has performed a significant creative or interpretive leap to generate its response.

    \item \textbf{Semantic Instability (The "Variance" Dimension):} The second dimension measures the semantic distances between prompts and responses across a range of semantically equivalent prompts. It quantifies how consistent the model's response strategy is, in terms of its semantic alignment with the prompt. We capture this using our practical score $\mathcal{S}_H$ which is given by a weighted average of the Ensemble JSD and the Wasserstein distances which are calculated over the entire distribution of generated sentences, making the score highly sensitive to the spread and consistency of the response cloud. Variability of answers is also factored into the construction of semantic basis by agglomerative clustering of prompt and response sentences. 
\end{enumerate}

By establishing threshold values, $KL_{\star}$ and $\mathcal{S}_{\star}$, we can classify any given response into one of four regimes based on these two dimensions. This provides two binary outputs:
\begin{itemize}
    \item Is KL(Answer $||$ Prompt) greater or less than $KL_{\star}$? This distinguishes between a \textit{low-exploration} and a \textit{high-exploration} response.
    \item Is the $\mathcal{S}_H$ score greater or less than $\mathcal{S}_{\star}$? This distinguishes between a \textit{low-variability} (stable/convergent) and a \textit{high-variability} (unstable/divergent) response.
\end{itemize}

This two-dimensional diagnostic is the core theoretical principle of our framework. It moves beyond a single pass/fail score and provides the information-theoretic foundation for our Semantic Box, a tool for classifying the nuanced behaviors of LLMs—from faithful recall to confident confabulations.

\subsection{From Theory to a Practical Hallucination Score}
The theoretical perspective above motivates a complexity-normalized indicator for hallucination. The entropy of a response, $H(P_r)$, naturally increases with prompt complexity. The latter can be proxied by the prompt entropy $H(P_q)$. A purely high-entropy response is not necessarily a hallucination. However, the portion of the response's entropy that is \textit{not} explained by its mutual information $ I(P_q; P_r) $ with the prompt can be a strong signal. This inspires the following idealized hallucination indicator:
\begin{equation}
\label{phi_theory_0}
\phi(q, r) = \frac{H(P_r) - I(P_q; P_r)}{H(P_q)} = \frac{H(P_r|R_q)}{H(P_q)}
\end{equation}
This ratio, to be referred to as the Normalized Condtional Entropy 
(NCE) in what follows, is given by the conditional entropy of the response given the prompt normalized by the prompt's own entropy, and captures the proportion of response complexity not justified by the prompt. High values of this metric may signal potential large deviations of the response from the topics found in the prompt, further proposing a potential for hallucinations in such responses. 

While Eq.(\ref{phi_theory_0}) provides a powerful theoretical blueprint, it is rigid. For a more flexible and empirically robust metric, below  we construct a more flexible practical hallucination score, $\mathcal{S}_{H}$, whose components are obtained from the same information-theoretical metrics, and also includes a distributional distance component. Details of our constructions and the exact definition of the the aggregate metric $S_H $ 
and the 'semantic exploration' metric KL[Answer$||$ Prompt] will be given below in a due course.
The remainder of this section details the process for computing the components of this practical score.

\subsection{Algorithmic Framework}
To compute the components of the $\mathcal{S}_H$ score, we follow a four-stage process: (1) data generation and embedding, (2) joint semantic clustering to identify shared topics, (3) computation of key information-theoretic and distributional distance metrics, and (4) aggregation into the final score. A high-level overview is presented in Algorithm \ref{alg:smi}.

\subsection{Data Generation and Joint Embedding}

To robustly capture the semantic space of both the input and output, we first generate a diverse set of prompt-answer pairs. For each initial prompt, we generate $M$ paraphrases. For each of these $M$ prompts, we then generate $N$ answers from the target LLM using temperature-based sampling.

The next step is to process all generated text at the sentence level. All sentences from both prompts and answers are segmented and then individually embedded into a common high-dimensional vector space. For all experiments in this paper, we utilize the \textbf{\texttt{Qwen3-Embedding-0.6B}} model, a pre-trained sentence-transformer, to perform this embedding. This process results in two sets of sentence embeddings: $\mathcal{P} = \{p_1, \dots, p_{S_P}\}$ for the prompt sentences and $\mathcal{A} = \{a_1, \dots, a_{S_A}\}$ for the answer sentences, which are then combined for the joint clustering analysis.

\subsection{Joint Semantic Clustering and Topic Estimation}

A key innovation of our method is the use of joint clustering. Rather than analyzing prompts and answers in isolation, we pool their sentence embeddings into a single dataset, $\mathcal{S} = \mathcal{P} \cup \mathcal{A}$. We then apply hierarchical agglomerative clustering with Ward's linkage to this combined set to identify shared semantic topics. This bottom-up approach builds a hierarchy of clusters without requiring the number of clusters to be specified beforehand, which is ideal for handling prompts of varying complexity. 
In our experiments, we however do not use this flexibility offered by agglomerative clustering algorithms, and instead fix the number of clusters by first running the K-means algorithm on our embeddings for different number of clusters. 
We then estimate the optimal number of clusters, $k$, by identifying the "elbow point" on the inertia plot, which represents the most statistically significant number of distinct semantic topics present across both prompts and answers. The number of clusters determined in this way is then passed to the hierarchical clustering algorithm as a constraint.\footnote{We also played with an alternative way of specifying the number of clusters in a hierarchical clustering by defining a distance threshold, and found similar results.} 
This joint clustering approach is crucial as it ensures that semantically similar sentences, regardless of whether they originate from a prompt or an answer, are grouped into the same topic cluster, providing a direct and meaningful basis for comparing their content distributions.

\subsection{Computing Topic-Based Alignment and Divergence Metrics}

With all sentences assigned to one of $k$ shared topic clusters, the final step is to quantify the semantic relationship between the prompt and response distributions. Our analysis framework computes a primary divergence score for our final hallucination score, alongside several supplementary metrics for comprehensive diagnostics.

\paragraph{Primary Alignment Metric: Jensen-Shannon Divergence.}
Our core measure of semantic alignment is the Jensen-Shannon Divergence (JSD), which quantifies the similarity between the topic distribution of the prompts, denoted by $\mathbf{P}$, and that of the answers, denoted by $\mathbf{A}$. The JSD is a symmetric and bounded version of the Kullback-Leibler (KL) divergence, making it a robust metric for comparing probability distributions. It is defined via a mixture distribution $\mathbf{M} = \frac{1}{2}(\mathbf{P} + \mathbf{A})$, as follows:
\begin{equation}
    D_{\text{JS}}(\mathbf{P} \| \mathbf{A}) = \frac{1}{2} \left( D_{\text{KL}}(\mathbf{P} \| \mathbf{M}) +  D_{\text{KL}}(\mathbf{A} \| \mathbf{M}) \right)
\end{equation}
The resulting score ranges from 0 (for identical distributions) to 1 (for maximally different distributions, using a base-2 logarithm). This direct comparison of the overall "topic mix" provides an intuitive measure of semantic drift. In our framework, we compute this metric using two distinct methodologies---a \textit{Global (Aggregate-First)} approach and an \textit{Ensemble (Average-Later)} approach---to capture different facets of the model's behavior, as detailed in the following subsection. The Ensemble JSD serves as a key component of our final $\mathcal{S}_H$ score.

\paragraph{Diagnostic Metric: Averaged Mutual Information and its Visualization.}
To diagnose the underlying dependency structure between prompt and answer topics, we compute the Mutual Information (MI) from an \textbf{Averaged Joint Probability Distribution}. This ensures that our numerical MI metric corresponds directly to our visualization.

Let $X$ be the discrete random variable representing the topic index of a sentence drawn from the prompt, and let $Y$ be the discrete random variable for the topic index of a sentence drawn from the corresponding answer.
The process for estimating their joint probability distribution is as follows:

\begin{enumerate}
    \item For each of the $M$ prompt-answer pairs $(P_m, A_m)$, we construct a local $k \times k$ contingency table, $\mathbf{C}_m$, based on sentence-level co-occurrence within that single pair.
    \item We normalize each local table to create a local joint probability distribution, $P_m(X,Y) = \mathbf{C}_m / \sum_{i,j}\mathbf{C}_m[i,j]$.
    \item The final, representative distribution is the element-wise average of these $M$ local distributions:
\end{enumerate}
\begin{equation}
\label{P_xy_theory}
    P_{\text{avg}}(X,Y) = \frac{1}{M} \sum_{m=1}^{M} P_m(X,Y)
\end{equation}
This averaged distribution is visualized as a heatmap to show the average dependency structure. The MI is then calculated directly from this matrix using the standard formula, providing a single, consistent measure of the average statistical dependence between prompt and answer topics.

\paragraph{Distributional Shift in Embedding Space.}
To measure the overall shift in semantic content, we go beyond simple topic distributions and compare the full geometry of the sentence embeddings. We compute the \textbf{1-Wasserstein distance}, also known as the Earth Mover's Distance, between the distribution of prompt embeddings and the distribution of answer embeddings.

Intuitively, the Wasserstein distance represents the minimum "cost" required to transform one distribution into the other, akin to finding the most efficient way to move a pile of earth from one configuration to another. In our context:
\begin{itemize}
    \item The two "piles of earth" are the clouds of prompt and answer sentence embeddings in the high-dimensional vector space.
    \item The "cost" of moving a unit of earth is the Euclidean distance between a prompt embedding and an answer embedding.
\end{itemize}
Unlike a simple comparison of centroids, the Wasserstein distance is sensitive to the shape, spread, and internal structure of the distributions.We emphasise here that the calculation of the  Wasserstein distance does not depend on our joint clustering amd cluster labeling of sentences from the prompt and response, as it instead operates on the original higher dimensional clouds of all prompts and all answers.

\subsection{Global vs Ensemble-based Information Metrics}

With sentences assigned to shared topic clusters, we compute a suite of metrics to provide a comprehensive view of the prompt-response relationship. We distinguish between two fundamental calculation methodologies: \textit{Global (Aggregate-First)} and \textit{Ensemble (Average-Later)}.

\paragraph{Global (Aggregate-First) Metrics.}
These metrics provide a macro-level view of the overall divergence between the entire corpus of prompt sentences and the entire corpus of answer sentences. They are calculated by first pooling all sentences to create two global topic distributions:
\begin{itemize}
    \item $P(X)$: The probability distribution of topics across \textit{all} sentences from the $M$ paraphrased prompts. The $i$-th component, $P(X=i)$, is the fraction of all prompt sentences assigned to cluster $C_i$.
    \item $P(Y)$: The probability distribution of topics across \textit{all} sentences from the $M \times N$ generated answers. The $j$-th component, $P(Y=j)$, is the fraction of all answer sentences assigned to cluster $C_j$.
\end{itemize}
From these two global distributions, we compute the \textbf{Global Jensen-Shannon Divergence ($D_{\text{JS}}$)}and the two directions of \textbf{Global Kullback-Leibler Divergence ($D_{\text{KL}}$)}.

\paragraph{Ensemble (Average-Later) Metrics.}
These metrics provide a more granular, micro-level view by measuring the \textit{average} behavior across the $M$ individual paraphrase experiments. This approach captures the pair-wise variability that can be lost in global aggregation. The process is as follows:
\begin{enumerate}
    \item For each of the $M$ prompt-answer pairs $(P_m, A_m)$ in the ensemble, we create a \textit{local} topic distribution for its prompts, $P_m(X)$, and a local distribution for its answers, $P_m(Y)$.
    \item We then compute the local divergence metrics for this single pair, e.g., $D_{\text{JS}}(P_m(X) || P_m(Y))$.
    \item The final Ensemble metric is the average of these $M$ local values. For example, the \textbf{Ensemble JSD} is:
\end{enumerate}
\begin{equation}
    D_{\text{JS}}^{\text{ens}} = \frac{1}{M} \sum_{m=1}^{M} D_{\text{JS}}(P_m(X) || P_m(Y))
\end{equation}
We compute the \textbf{Ensemble KL Divergences} in the same manner. This "average-later" methodology is also used to calculate our robust diagnostic, the \textbf{Ensemble Mutual Information}, via the identity $I^{\text{ens}}(X;Y) = H(Y) - \frac{1}{M} \sum_{m=1}^{M} H(Y_m|X_m)$, ensuring methodological consistency across all our fine-grained diagnostic measures, as specified in more details in the next subsection. These Ensemble metrics are reported in our results tables to provide a deeper understanding of the model's behavior on an instance-by-instance basis.

\subsection{Ensemble Mutual Information (EMI).}

While the contingency table can be used to estimate MI according to 
Eq.(\ref{P_xy_theory}), we also consider a different way to compute MI via the identity $I(X;Y) = H(Y) - H(Y|X)$, where the conditional entropy is averaged over the ensemble of $M$ paraphrase experiments.
This MI metric calculated in this way will be referred to as 
the Ensemble Mutual Information (EMI). The EMI metric is designed to provide a more robust estimate of statistical dependence than a single, global calculation.

The implementation of the EMI metric involves two main steps:

\begin{enumerate}
    \item \textbf{Overall Answer Entropy ($H(Y)$):} This term represents the total uncertainty of the response topics. It is calculated using the standard Shannon entropy formula on the aggregate topic distribution of \textit{all} answer sentences, $P(Y)$.

    \item \textbf{Average Conditional Entropy ($H(Y|X)$):} This term represents the remaining uncertainty in the answer topics after the prompt topics are known. To compute this, we average the conditional entropy across the ensemble of $M$ experiments:
    \begin{itemize}
        \item For each of the $M$ pairs $(P_m, A_m)$ in the ensemble, we first construct a \textit{local} $k \times k$ contingency table, $\mathbf{C}_m$, based on sentence-level co-occurrence.
        \item From this local table, we calculate the local conditional entropy, $H(Y_m|X_m)$, which measures the answer uncertainty for that specific instance.
        \item The final conditional entropy is the average of these local values, representing the expected remaining uncertainty:
    \end{itemize}
\end{enumerate}
\begin{equation}
    H(Y|X) = \frac{1}{M} \sum_{m=1}^{M} H(Y_m|X_m)
\end{equation}
The \textbf{Ensemble Mutual Information (EMI)} is then the difference between these two entropy terms: $I(X;Y) = H(Y) -  H(Y|X)$. This value, which quantifies the average reduction in uncertainty across the experimental ensemble, is reported in our results tables (e.g., Table~\ref{tab:smi_gradient_results}).

\subsection{SDM Metrics: Normalized Conditional Entropy, $S_H $ and KL Scores}

Our framework produces three key indicators (Semantic Diverge Metrics) for analysis: 
Normalized Conditional Entropy $\phi$, the aggregate SDM Hallucination Score $\mathcal{S}_H$, and KL divergence KL(Answer $ || $ Prompt).


\paragraph{Normalized Conditional Entropy $\phi(Y|X)$.}
This theoretical score quantifies the proportion of the answer's complexity that is \textit{not} explained by its shared information with the prompt, see Eq.(\ref{phi_theory_0}):
\begin{equation}
    \phi(Y|X) = \frac{H(Y|X)}{H(X)} = \frac{H(Y) - I(X;Y)}{H(X)}
\end{equation}
A value near 1.0 suggests appropriate response complexity, while a value significantly greater than 1.0 signals a potential hallucination due to unexplained complexity.

\subsection{The Practical SDM Hallucination Score $\mathcal{S}_H$}

While our framework produces a rich set of diagnostic metrics, a single, practical score is desirable for ranking response stability. We construct the $\mathcal{S}_H$ score as a weighted sum of our two most robust and informative components: the Ensemble Jensen-Shannon Divergence and the Wasserstein Distance.

Importantly, these two metrics measure fundamentally different aspects of the "distance" between the prompts and responses. They operate on different representations of the data and are therefore complementary:
\begin{itemize}
    \item The \textbf{JSD metric} operates on the \textit{low-dimensional, discretized topic space} created by our joint clustering. It measures the divergence between the categorical distributions of topics, telling us if the \textit{thematic composition} of the answers has shifted relative to the prompts.
    \item The \textbf{Wasserstein distance}, in contrast, operates directly on the \textit{high-dimensional, continuous embedding space}. It measures the geometric distance between the raw "clouds" of sentence embeddings, completely independent of any clustering. It tells us the overall shift in the \textit{semantic content and meaning}, even if the high-level topics remain the same.
\end{itemize}
By combining these two measures, our framework captures both high-level thematic drift and lower-level shifts in semantic content.

The two final components for our score are:
\begin{enumerate}
    \item \textbf{Global Distributional Shift ($W_d$):} The 1-Wasserstein distance between the full distributions of all prompt and answer embeddings.
    \item \textbf{Ensemble Semantic Divergence ($D^{\text{ens}}_{\text{JS}}$):} The Ensemble Jensen-Shannon Divergence, our most sensitive measure of topic alignment.
\end{enumerate}

These components are combined into the final score defined as their weighted sum, scaled by the entropy of the input:
\begin{equation}
    \mathcal{S}_{H} = \frac{ w_{\text{wass}} \cdot W_d + w_{\text{jsd}} \cdot D^{\text{ens}}_{\text{JS}} }{H(P)}
    \label{eq:sh_score_final}
\end{equation}
We scale our final score by the prompt entropy $ H(P) $ in order to partially reduce the dependence of our semantic divergence metric on the complexity of the prompt itself, and hence to more reliable compare results obtained across different types of prompts.

The choice of weights, $w_{\text{jsd}}$ and $w_{\text{wass}}$, is non-trivial, as the two metrics operate on different intrinsic scales. A naive weighting would be arbitrarily dominated by the metric with the larger average magnitude. To create a more balanced and meaningful score, we calibrated the weights empirically with the goal of having the Ensemble JSD contribute approximately 60\% to the final score's value on average, and the Wasserstein Distance the remaining 40\%.

As detailed in the Experiments section (Section~\ref{sec:experiments}), we performed this calibration process separately for our two distinct experimental sets, which represent very different types of generative tasks. This analysis revealed a critical insight: the optimal weights derived for both sets were nearly identical. This consistency suggests that the relative scaling between these two metrics is stable across different task types, justifying the use of a single, unified set of weights. For all experiments in this paper, we therefore use the final calibrated weights $w_{\text{jsd}}=0.7$ and $w_{\text{wass}}=0.3$. This score, which combines our most robust metrics in a balanced and empirically grounded manner, serves as our primary practical tool for ranking response stability.

\subsection{The KL score}

In addition to confirming usefulness of the $ \Phi $ and $ S_H $ metrics, our experiments described below identified a very important complimentary metric which is the KL divergence KL(Answer$||$Prompt). Unlike $ S_H$ which is a bi-directional symmetric metric, the KL divergence is asymmetric, reflecting the flow of semantic information from the prompt to the LLM response.
This metric turns out to be very sensitive to the amount of semantic generative exploration assumed in the prompt, and contained in the LLM's response given the semantic richness of the prompt. Our experiments suggest that this metric should be computed and monitored alongside the $ \Phi $ and $ S_H $ metrics for a better overall assessment of semantic coherence and stability of a given prompt-response pair.
Similarly to our definition of metric $ S_H $, we additionally scale our final metric by the entropy of the prompt $ H[P] $, so our final KL score is
\begin{equation}
	KL(Answer || Prompt) = \frac{D_{KL}(Answer || Prompt)}{
	H(Prompt)}
\end{equation}

\begin{algorithm}[h!]
\caption{Semantic Divergence Metrics (SDM) for Stability Analysis}
\label{alg:smi}
\begin{algorithmic}[1]
\Require Original prompt $Q$, Target LLM $\mathcal{L}$, Sentence Encoder $\mathcal{E}$
\State Generate $M$ paraphrases $\{Q_m\}$ of $Q$ and $N$ answers $\{A_{m,n}\}$ for each.
\State Segment all prompts and answers into sentences.
\State Embed all sentences using $\mathcal{E} \rightarrow \mathcal{P}$ (prompt embeddings), $\mathcal{A}$ (answer embeddings).
\State Combine embeddings: $\mathcal{S} \leftarrow \mathcal{P} \cup \mathcal{A}$.
\State Determine optimal cluster count $k$ on $\mathcal{S}$ using the Elbow Method with K-means clustering
\State Perform hierarchical clustering on $\mathcal{S}$ with $k$ clusters $\rightarrow \{C_1, \dots, C_k\}$.
\State Get cluster labels for all prompt sentences ($L_\mathcal{P}$) and answer sentences ($L_\mathcal{A}$).

\Statex \textit{// --- Compute Global (Aggregate-First) Metrics ---}
\State Derive global topic distributions $\mathbf{P}_{\text{global}}$ from $L_\mathcal{P}$ and $\mathbf{A}_{\text{global}}$ from $L_\mathcal{A}$.
\State Compute Global $D_{\text{JS}}$, Global $D_{\text{KL}}$ from these distributions.
\State Compute the 1-Wasserstein distance $W_d$ between the full embedding sets $\mathcal{P}$ and $\mathcal{A}$.

\Statex \textit{// --- Compute Ensemble (Average-Later) Metrics ---}
\ForAll{each of the $M$ prompt-answer pairs $(P_m, A_m)$}
    \State Derive local topic distributions $\mathbf{P}_m$ and $\mathbf{A}_m$ for the current pair.
    \State Calculate local divergences, such as $D_{\text{JS}}(\mathbf{P}_m \| \mathbf{A}_m)$ and $D_{\text{KL}}(\mathbf{A}_m \| \mathbf{P}_m)$.
\EndFor
\State Compute final Ensemble metrics (e.g., $D^{\text{ens}}_{\text{JS}}$, $D^{\text{ens}}_{\text{KL}}$, the NCE score $ \Phi $, etc.) by averaging the local values.

\Statex \textit{// --- Calculate Final Score ---}
\State Calculate the final Hallucination Score: $\mathcal{S}_{H} \leftarrow \left( w_{\text{jsd}} \cdot D^{\text{ens}}_{\text{JS}} + w_{\text{wass}} \cdot W_d \right)/ H(P) $.
\State \Return Key metrics: $\mathcal{S}_{H}$, $ \Phi $, $D_{\text{KL}}(\mathbf{A} \| \mathbf{P})$
\end{algorithmic}
\end{algorithm}

\section{Computational Complexity Analysis}

The computational cost of our SDM framework is an important consideration for practical deployment. We analyze its time and space complexity below.

\subsection{Time Complexity}
The overall complexity is dominated by three main stages: embedding, clustering, and the calculation of divergence metrics. Let $S$ be the total number of sentences (prompts and answers combined), $M$ be the number of paraphrased prompts, $k$ be the number of topic clusters, and $d$ be the embedding dimension.

\begin{itemize}
    \item \textbf{Sentence Embedding:} This step is linear in the total number of sentences, with a complexity of $O(S \cdot d)$.
    
    \item \textbf{Hierarchical Clustering:} This is the most computationally expensive step. Standard agglomerative clustering has a time complexity of $O(S^2 \log S)$ or $O(S^2)$ in efficient implementations for a fixed number of clusters, which dominates the process for large $S$.
    
    \item \textbf{Metric Calculation:}
    \begin{itemize}
        \item The \textit{Global} divergence metrics (JSD, KL) are highly efficient, requiring a single pass over the $S$ sentence labels to build distributions and then a $O(k)$ operation.
        \item The \textit{Ensemble} divergence metrics require iterating through the $M$ pairs. Inside the loop, building local distributions takes time proportional to the number of sentences in each pair. The overall complexity is approximately $O(S)$.
        \item The \textit{Wasserstein distance} is computed on the full sets of prompt and answer embeddings ($S_P$ and $S_A$ sentences, respectively). Its exact complexity is roughly $O((S_P+S_A)^3 \log(S_P+S_A))$, but this is typically fast as the number of sentences per experiment is manageable.
    \end{itemize}
\end{itemize}
Therefore, the total time complexity is primarily driven by the hierarchical clustering step, $O(S^2)$. This makes the framework practical for typical scenarios where the total number of sentences $S$ is in the hundreds.

\subsection{Space Complexity}
The primary memory requirement comes from storing the full set of sentence embeddings, which has a space complexity of $O(S \cdot d)$. The cost matrix for the Wasserstein distance requires $O(S_P \cdot S_A)$ space during its calculation. All other data structures, such as the topic distributions and co-occurrence matrices, are small, requiring only $O(k^2)$ or $O(k)$ space. This makes the method memory-efficient and scalable with respect to the number of topics.

\section{Experiments}
\label{sec:experiments}

To validate our framework's capabilities, we conducted two distinct sets of experiments using the \texttt{gpt-4o} model. The first set was specifically designed to test the framework's sensitivity across a controlled "stability gradient," from highly factual to purely creative tasks. The second set used a more diverse range of prompt types to identify different modes of model behavior, including a "confident hallucination" state. For all experiments, we generated 10 paraphrases and 4 responses per paraphrase.

\subsection{Experiment Set A: Controlled Stability Gradient}

\subsubsection{Prompt Descriptions}
This set included three prompts of comparable length ($\approx 150$ words) but with increasing degrees of creative freedom, designed to elicit responses with a clear gradient of stability.
\begin{enumerate}
    \item \textbf{High Stability (Hubble):} A detailed, fact-based query asking for a summary of the Hubble Space Telescope. We hypothesized this would produce the lowest scores, indicating minimal uncertainty and high stability.
    
    \item \textbf{Moderate Stability (Hamlet):} An interpretive summary task based on a factual source (Shakespeare's 'Hamlet'), structured into three specific thematic paragraphs. We hypothesized this would produce mid-range scores, reflecting the allowed variance in summarization and interpretation.
    
    \item \textbf{Low Stability (AGI Dilemma):} A highly speculative and ungrounded prompt requiring the model to create a fictional ethical scenario. We hypothesized this would produce the highest scores, indicating significant semantic drift and high uncertainty inherent in a creative task.
\end{enumerate}

The full set of prompts is presented in Appendix~\ref{sec:appendix_prompts_v2}.

\subsubsection{Quantitative Analysis and Discussion}

The results, shown in Table~\ref{tab:smi_gradient_results}, clearly demonstrate the framework's ability to track the stability gradient and provide a multi-faceted view of the model's behavior. The key findings are as follows:

\begin{itemize}

    \item \textbf{The SDM Score Tracks the Stability Gradient.} As hypothesized, the final, complexity-scaled SDM score $ S_H $  increases monotonically with the prompt's ambiguity: $0.2918 \rightarrow 0.3297  \rightarrow 0.5919$. This confirms that our final score, normalized by the prompt's intrinsic complexity, effectively quantifies the relative semantic instability of the LLM's responses. The NCE indicator, $\Phi$, also follows this trend.

    \item \textbf{Component Metrics Reveal the Nature of Instability.} By decomposing the $ S_H $ score, we can understand the source of the instability. The Low Stability prompt's high score is driven primarily by its large \textbf{Ensemble JSD} (0.6626). Conversely, the Moderate Stability prompt's score is characterized by the highest absolute \textbf{Wasserstein Distance} (0.8782) in the set, suggesting that while the model's topic structure was somewhat constrained, the specific semantic content it generated varied significantly—the signature of an interpretive task.

    \item \textbf{KL Divergence KL(Answer $||$ Prompt) as a Measure of Semantic Exploration.} The Ensemble KL(Answer $||$ Prompt) provides the most dramatic signal of generative behavior. For the highly creative `AGI Dilemma` prompt, the value explodes to 19.5591, confirming that the model must invent a vast amount of new semantic content to answer. Interestingly, the `Hubble` prompt (7.1488) shows a higher KL divergence than the `Hamlet` prompt (5.1408). This suggests that synthesizing a broad range of disparate facts for the Hubble summary required more semantic exploration than adhering to the tight, three-part thematic structure of the Hamlet summary.

    \item \textbf{Interpreting the Diagnostic MI Metrics.} The diagnostic metrics suggest that standard MI measures may not be well-suited for this task. The `Averaged MI` remained near-zero for all prompts. However, the `Ensemble MI` for the Moderate Stability prompt (0.1490 bits) was an order of magnitude higher than for the other two. This hints that the sentence-level dependencies in the structured, interpretive 'Hamlet' task were more complex and predictable than in both the rigid factual recall and the unconstrained creative generation tasks, a subtle insight missed by other metrics.

    \item \textbf{Limitations of Semantic Entropy.} The baseline Semantic Entropy (SE) metrics fail to track the stability gradient. The `SE (Original Prompt Only)` is highest for the factual 'Hubble' prompt (2.2190) and lowest for the interpretive 'Hamlet' prompt (0.8524). This counter-intuitive result demonstrates the core weakness of prompt-agnostic methods, which incorrectly flag a diverse, complex, but correct answer as potentially problematic.
\end{itemize}

\begin{table}[h!]
\centering
\caption{Summary of Results for Experiment Set A. Metrics ($ S_H $) and Ensemble KL(A $ || $ P) are normalized by the Global Prompt Entropy $H(P)$ for cross-prompt comparability.}
\label{tab:smi_gradient_results}
\resizebox{\textwidth}{!}{%
\begin{tabular}{lccc}
\toprule
\textbf{Metric} & \textbf{High Stability} & \textbf{Moderate Stability} & \textbf{Low Stability} \\
& (Hubble) & (Hamlet) & (AGI Dilemma) \\
\midrule
\textbf{SDM Score $S_H $} & \textbf{0.2918} & \textbf{0.3297} & \textbf{0.5919} \\
\textbf{Norm. Cond. Entropy $\Phi$} & 0.9489 & 1.0507 & 1.5074 \\
\midrule
\textit{Global Divergence Metrics} & & & \\
Global Prompt Entropy H(P) & 1.9165 & 1.8295 & 1.2147 \\
Global JSD & 0.3337 & 0.4451 & 0.6205 \\
Global KL(P $||$ A) & 0.4185 & 0.7629 & 1.4513 \\
Global KL(A $||$ P) & 0.5241 & 9.1586 & 11.3269 \\
Entropy Difference H(A) - H(P) & 0.0849 & 0.0720 & 0.6013 \\
\midrule
\textit{Ensemble Divergence Metrics} & & & \\
Ensemble JSD & 0.4492 & 0.4854 & 0.6626 \\
Ensemble KL(A $||$ P) & 7.1488 & 5.1408 & 19.5591 \\
\midrule
\textit{Other Metrics} & & & \\
Wasserstein Distance & 0.8162 & 0.8782 & 0.8503 \\
Ensemble MI (bits) & 0.0174 & 0.1490 & 0.0113 \\
Averaged MI (bits) & 0.0023 & 0.0047 & 0.0013 \\
\midrule
\textit{Semantic Entropy Baseline} & & & \\
SE (Original Prompt Only) & 2.2190 & 0.8524 & 1.9491 \\
Mean SE (Across Paraphrases) & 1.5899 & 1.8952 & 1.3708 \\
\bottomrule
\end{tabular}%
}
\end{table}

\subsubsection{Visual Analysis of Topic Co-occurrence}
\label{sec:experiments_viz}

\begin{figure}[h!]
    \centering
    
    \begin{subfigure}[b]{0.33\textwidth} 
        \centering
        \includegraphics[width=\textwidth]{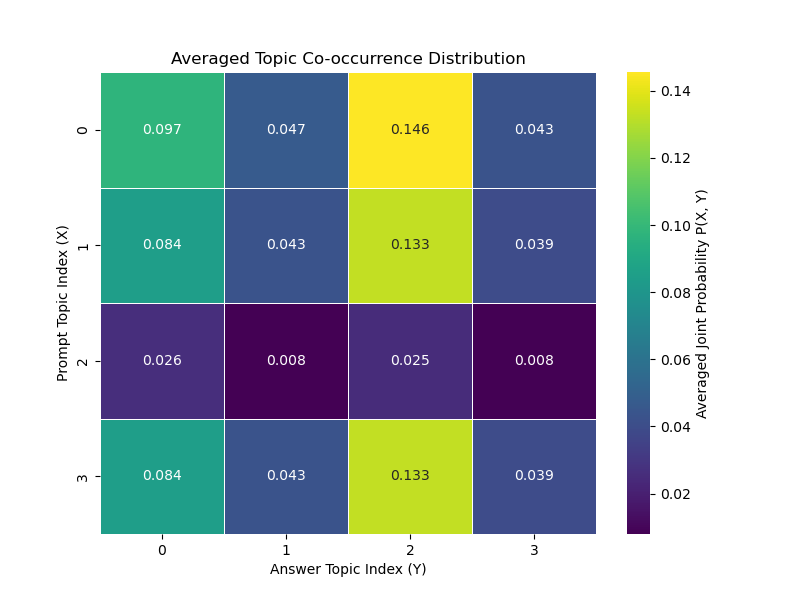}
        \subcaption{High Stability Prompt}
        \label{fig:high_stability_heatmap}
    \end{subfigure}%
    \hfill 
    \begin{subfigure}[b]{0.33\textwidth} 
        \centering
        \includegraphics[width=\textwidth]{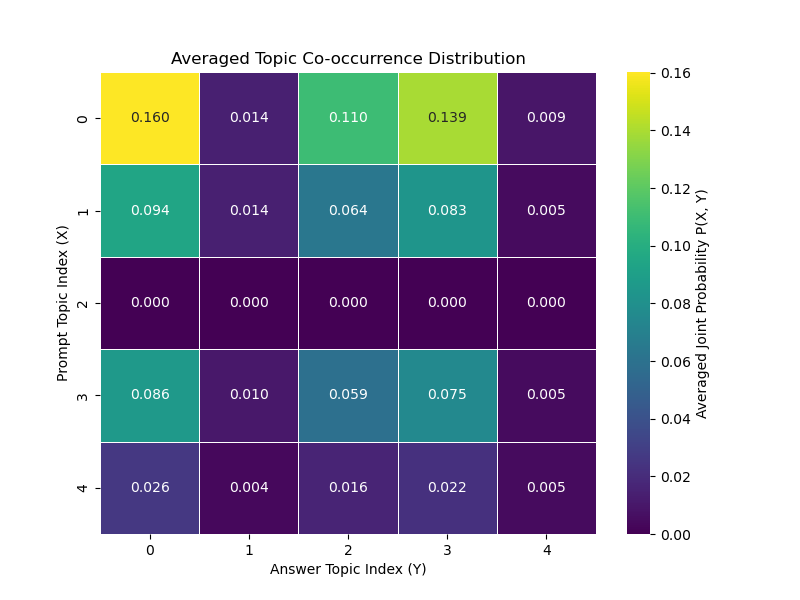}
        \subcaption{Moderate Stability Prompt}
        \label{fig:moderate_stability_heatmap}
    \end{subfigure}%
    \hfill
    \begin{subfigure}[b]{0.33\textwidth} 
        \centering
        \includegraphics[width=\textwidth]{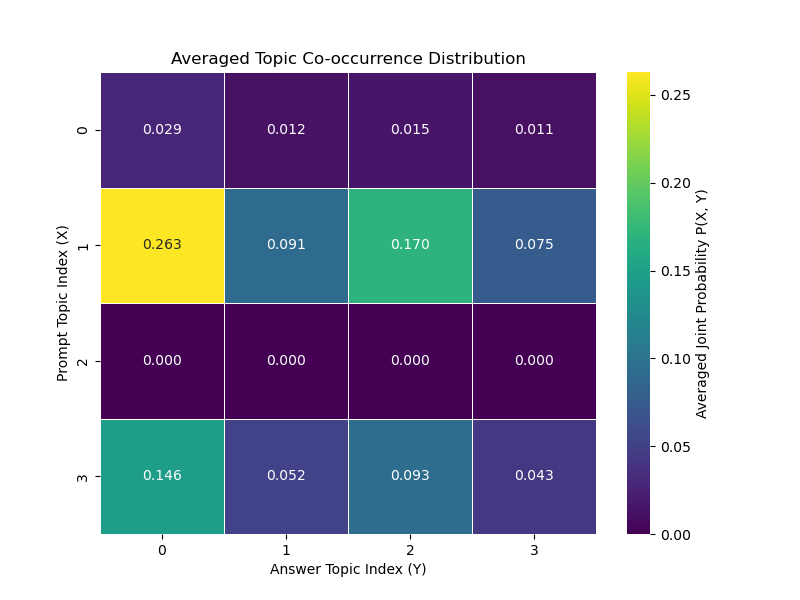}
        \subcaption{Low Stability Prompt}
        \label{fig:low_stability_heatmap}
    \end{subfigure}
    
    \caption{Averaged Topic Co-occurrence Distributions for Experiment Set A. The heatmaps show the averaged joint probability $P_{\text{avg}}(X,Y)$ between prompt topics (Y-axis) and answer topics (X-axis) for the three prompts in the stability gradient experiment.}
    \label{fig:topic_cooccurrence_A}
\end{figure}

The averaged topic co-occurrence distributions, visualized as heatmaps in Figure~\ref{fig:topic_cooccurrence_A}, provide a powerful illustration of the model's response strategy across the three stability scenarios. By first analyzing the distribution of answer topics (X-axis) and then its consistency across prompt topics (Y-axis), we can visually identify the signatures of factual recall, structured interpretation, and creative generation.

\begin{itemize}
    \item \textbf{High Stability (Hubble):} The heatmap (a) for the factual `Hubble` prompt shows a clear and focused response strategy. The answer distribution (X-axis) is strongly unimodal, with probability mass heavily concentrated on a single primary answer topic (topic 2, with probabilities of 0.146 and 0.133). However, this response is \textbf{conditional} on the prompt's phrasing. The model produces this focused answer for prompt topics 0, 1, and 3, but the mapping is much weaker for prompt topic 2. This suggests a stable but nuanced recall, where the model has a preferred way of answering but is sensitive to variations in the query.

    \item \textbf{Moderate Stability (Hamlet):} The `Hamlet` heatmap (b) reveals a more complex, multi-modal answer strategy, with the model utilizing several distinct topics in its responses (notably 0, 2, and 3). Critically, the response is highly \textbf{non-uniform} across the Y-axis. The mapping is strongly conditional: prompt topic 0 maps most strongly to answer topic 0 (0.160), while prompt topic 2 produces no response at all (a row of zeros). This asymmetric, conditional mapping, where different prompt phrasings elicit distinctly different answer structures, is the clear visual signature of a structured interpretive task.

    \item \textbf{Low Stability (AGI Dilemma):} The heatmap (c) for the creative `AGI Dilemma` prompt displays the most extreme conditional behavior. The model's response strategy is highly convergent on a single answer topic (topic 0), but this response is triggered almost exclusively by a single prompt topic (topic 1), resulting in the brightest cell in any heatmap (0.263). Other prompt topics fail to elicit this strong creative response, and one (topic 2) produces no response at all. This "brittle" or "spiky" mapping suggests that for this highly abstract task, the model has found one specific interpretation that it can confidently elaborate on, but struggles to generate creative content for other phrasings of the same abstract prompt.
\end{itemize}

Ultimately, these heatmaps serve as a powerful visual diagnostic. They confirm the quantitative findings and provide an intuitive understanding of how the model's response strategy shifts from a stable but conditional recall, to a complex and multi-faceted interpretation, and finally to a highly focused but conditionally triggered creative generation.

\subsection{Experiment Set B: Diverse Prompt Types}


The prompts were designed to span a spectrum from highly factual to deliberately nonsensical:
\begin{enumerate}
    \item \textbf{Factual (Hubble):} A request for a concise, fact-based summary of the Hubble Space Telescope. This prompt is grounded in well-established public knowledge and is expected to produce stable, low-hallucination responses with a moderate number of distinct topics (history, capabilities, discoveries).
    
    \item \textbf{Complex Comparison (Keynes vs. Hayek):} A comparative analysis of two opposing economic theories. This prompt requires nuance and the ability to synthesize multiple complex concepts. While grounded in fact, it allows for more interpretive freedom and is expected to have a higher intrinsic complexity.
    
    \item \textbf{Forecasting (AI Trends):} A prompt asking for a summary of current AI trends and a prediction of their future impact. This task is inherently speculative and has no single correct answer, making it susceptible to generating plausible-sounding but unsubstantiated claims. It is expected to exhibit higher uncertainty.
    
    \item \textbf{Forced Hallucination (QCD \& Baroque Music):} A deliberately nonsensical prompt forcing the model to fabricate a connection between Quantum Chromodynamics and Baroque music. This is designed to be a stress test, where any coherent-sounding answer is, by definition, a hallucination.
\end{enumerate}

The full set of prompts is presented in Appendix~\ref{sec:appendix_prompts}.

\subsection{Quantitative Analysis of Experiment Set B}

The key metrics are summarized in Table~\ref{tab:smi_results_initial}. The experimental results provide a nuanced view of the model's response stability and reveal critical insights into the nature of hallucination and the function of our framework.

%

\begin{table}[h!]
\centering
\caption{Summary of Results for Experiment Set B. Metrics $ S_H $ and Ensemble KL(A $ ||$ P)  are normalized by the Global Prompt Entropy $H(P)$ for cross-prompt comparability.}
\label{tab:smi_results_initial}
\resizebox{\textwidth}{!}{%
\begin{tabular}{lcccc}
\toprule
\textbf{Metric} & \textbf{Factual} & \textbf{Complex Comp.} & \textbf{Forecasting} & \textbf{Forced Hallucination} \\
& (Hubble) & (Keynes/Hayek) & (AI Trends) & (QCD/Baroque) \\
\midrule
\textbf{SDM Score $ S_H$} & \textbf{0.1945} & \textbf{0.1419} & \textbf{0.1600} & \textbf{0.1100} \\
\textbf{Norm. Cond. Entropy $\Phi$} & 1.0142 & 1.0297 & 1.0040 & 0.9906 \\
\midrule
\textit{Global Divergence Metrics} & & & & \\
Global Prompt Entropy H(P) & 1.8674 & 2.2480 & 1.9183 & 2.5850 \\
Global JSD & 0.1421 & 0.1024 & 0.1140 & 0.0774 \\
Global KL(P $||$ A) & 0.0794 & 0.0435 & 0.0518 & 0.0237 \\
Global KL(A $||$ P) & 0.0842 & 0.0407 & 0.0527 & 0.0244 \\
Entropy Difference H(A) - H(P) & 0.0427 & 0.0650 & 0.0077 & 0.0244 \\
\midrule
\textit{Ensemble Divergence Metrics} & & & & \\
Ensemble JSD & 0.2330 & 0.1681 & 0.1203 & 0.0942 \\
Ensemble KL(A $||$ P) & 1.7206 & 0.8644 & 0.0334 & 0.0154 \\
\midrule
\textit{Other Metrics} & & & & \\
Wasserstein Distance & 0.6668 & 0.6708 & 0.7422 & 0.7276 \\
Ensemble MI (bits) & 0.0017 & 0.0178 & 0.0116 & 0.0155 \\
Averaged MI (bits) & 0.0001 & 0.0004 & 0.0000 & 0.0000 \\
\midrule
\textit{Semantic Entropy Baseline} & & & & \\
SE (Original Prompt Only) & 1.9428 & 2.2956 & 2.2450 & 0.6253 \\
Mean SE (Across Paraphrases) & 2.2293 & 2.3250 & 2.1229 & 1.6925 \\
\bottomrule
\end{tabular}%
}
\end{table}

\begin{itemize}
    \item \textbf{The Signal of "Confident Hallucination."} The most striking and important result is that the \textit{Forced Hallucination} prompt yielded the \textit{lowest} SDM score ($\mathcal{S}_H = 0.11$). This is counter-intuitive but reveals a crucial lesson: {\bf our framework measures semantic instability, not factual correctness}. The model, when faced with a nonsensical query, did not produce random or unstable outputs. Instead, it converged on a highly stable and consistent "evasion strategy." Because this fabricated response was semantically stable, its topic and content drift were minimal, resulting in the lowest score. This demonstrates that a very low SDM score on a known nonsensical prompt can be a powerful signal of a model's tendency to generate confident and consistent falsehoods rather than express uncertainty.

    \item \textbf{High Stability Across Different Task Types.} The three valid prompts all produced similarly low SDM scores (ranging from 0.1419 to 0.1945), indicating that the LLM generated highly stable and semantically aligned responses for all of them. This suggests that the model is equally stable when recalling facts, comparing complex ideas, and making structured predictions.The Wasserstein Distance was highest for the `AI Trends` prompt (0.7422), correctly capturing that the raw semantic content of the speculative answers drifted more than for the fact-based prompts.

    \item \textbf{Limitations of the Semantic Entropy Baseline.} The baseline SE metrics fail to provide a clear signal. For the `Forced Hallucination` prompt, the `SE (Original Prompt Only)` is extremely low (0.6253), which correctly identifies the low diversity of the "evasion" answers. However, for the three valid prompts, the SE scores are all high and do not clearly distinguish between them. This highlights that while low SE can indicate response consistency, high SE is an ambiguous signal, failing to differentiate between a healthy, diverse answer to a complex prompt and an unstable, hallucinated one. Our SDM framework resolves this ambiguity by normalizing for prompt complexity.

    \item \textbf{Interpreting the Diagnostic MI Metrics.} The `Ensemble MI` metric shows some variation across prompts, with the `Complex Comparison` prompt registering the highest value (0.0178 bits), hinting at a more structured sentence-level relationship in its interpretive responses.
    
  \item \textbf{KL Divergence as a Signal of Semantic Exploration.} While the KL divergence values in this set are uniformly low in comparison to the "generative scaffolding" prompts of Experiment Set A, a subtle but important pattern emerges. The `Factual` and `Complex Comparison` prompts exhibit an `Ensemble KL(Answer $||$ Prompt)` (1.7206 and 0.8644, respectively) that is orders of magnitude higher than for the `Forecasting` and `Forced Hallucination` prompts (0.0334 and 0.0154). This suggests that even for these contained-concept tasks, the model must perform a degree of semantic extrapolation to synthesize facts or compare nuanced economic theories, even though at a lesser scale than for prompts in Experiment Set A. 

   In contrast, the extremely low KL for the `Forecasting` and `Forced Hallucination` tasks indicates that the model is falling back on a highly repetitive, template-like response that requires almost no novel semantic exploration. This reinforces the finding that Ensemble KL(Answer $||$ Prompt) is a sensitive measure of the cognitive work of interpretation and synthesis.
\end{itemize}

In conclusion, these experiments demonstrate that the SDM score is not an absolute measure of truth, but a powerful, context-aware tool for quantifying the stability of an LLM's response. A high score suggests semantic drift and potential hallucination, while a low score on a well-posed prompt suggests a stable and correct answer. Critically, a low score on an ill-posed prompt can indicate a different kind of failure mode: a stable and confident hallucination.

\subsubsection{Visual Analysis of Topic Co-occurrence}

\begin{figure}[h!]
    \centering
    
    \begin{subfigure}[b]{0.48\textwidth}
        \centering
        \includegraphics[width=\textwidth]{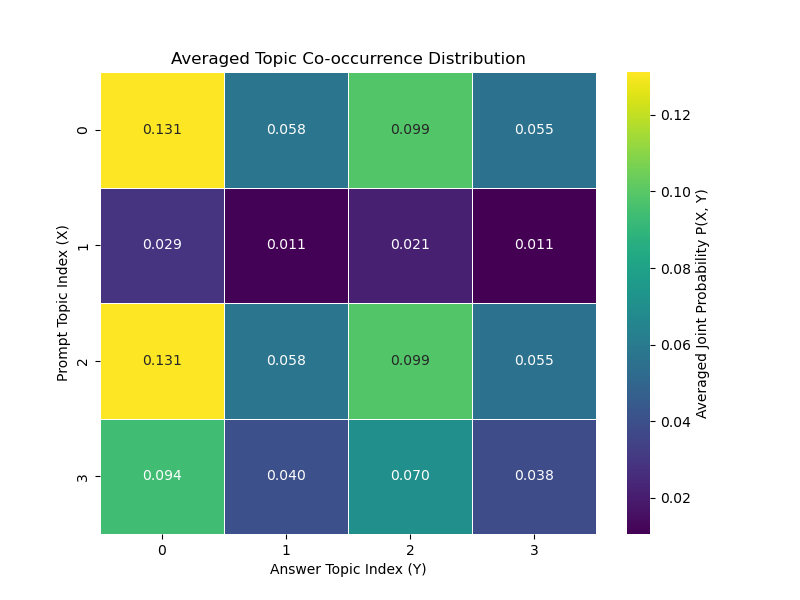}
        \subcaption{Factual (Hubble)}
        \label{fig:hubble_b_heatmap}
    \end{subfigure}
    \hfill 
    \begin{subfigure}[b]{0.48\textwidth}
        \centering
        \includegraphics[width=\textwidth]{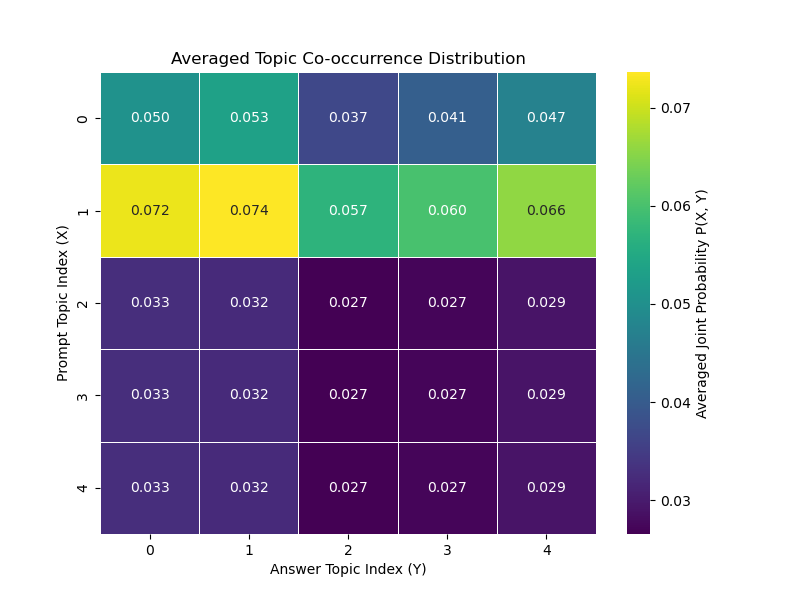}
        \subcaption{Complex Comparison (Keynes/Hayek)}
        \label{fig:keynes_b_heatmap}
    \end{subfigure}
    
    
    \begin{subfigure}[b]{0.48\textwidth}
        \centering
        \includegraphics[width=\textwidth]{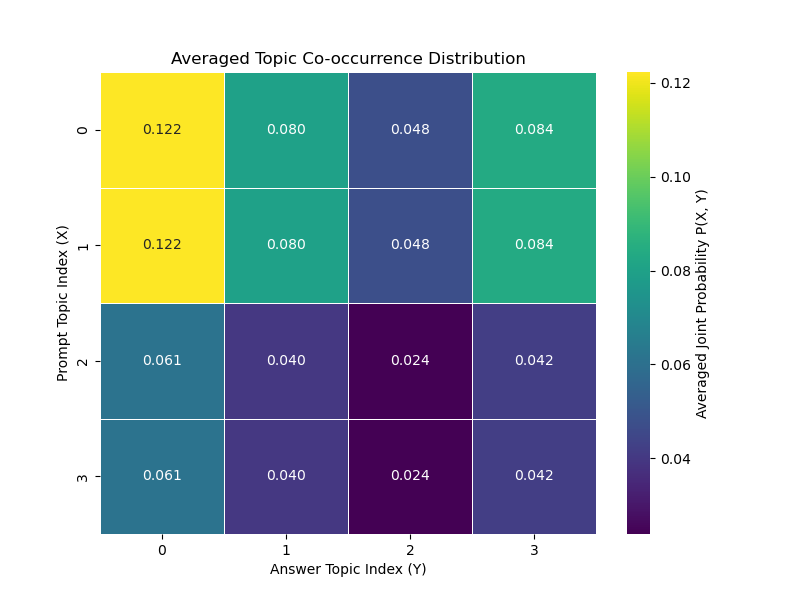}
        \subcaption{Forecasting (AI Trends)}
        \label{fig:forecast_b_heatmap}
    \end{subfigure}
    \hfill 
    \begin{subfigure}[b]{0.48\textwidth}
        \centering
        \includegraphics[width=\textwidth]{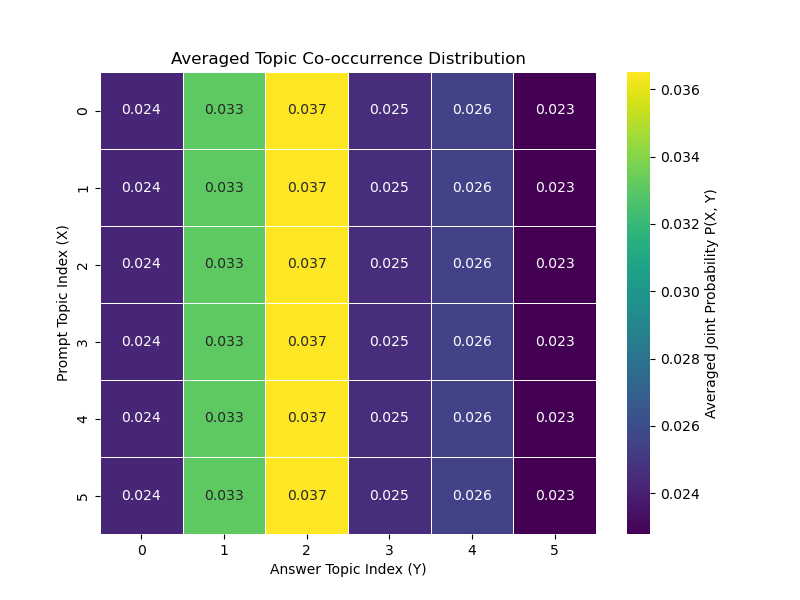}
        \subcaption{Forced Hallucination (QCD/Baroque)}
        \label{fig:qcd_b_heatmap}
    \end{subfigure}
    
    \caption{Averaged Topic Co-occurrence Distributions for Experiment Set B. The heatmaps show the averaged joint probability $P_{\text{avg}}(X,Y)$ between prompt topics (Y-axis) and answer topics (X-axis) for the four diverse prompt types.}
    \label{fig:topic_cooccurrence_B}
\end{figure}

The averaged topic co-occurrence distributions for Experiment Set B, visualized as heatmaps in Figure~\ref{fig:topic_cooccurrence_B}, reveal the distinct and stable response strategies the model employs for each task.

\begin{itemize}
    \item \textbf{Factual (Hubble):} The heatmap (a) displays a highly structured and \textbf{redundant mapping}. The probability distribution for prompt topic 0 is nearly identical to that for prompt topic 2, indicating that different phrasings of the factual query converge on the exact same bimodal answer distribution (a strong focus on answer topics 0 and 2). This visual redundancy is the signature of a stable, confident recall of a fixed set of facts.

    \item \textbf{Complex Comparison (Keynes/Hayek):} In contrast, the heatmap (b) for this prompt shows a \textbf{diffuse but structured distribution}. The probability mass is spread more broadly, with a noticeable concentration in the block defined by prompt topic 1 and answer topics 0 and 1. This less-focused pattern suggests the model is correctly engaging with a more complex topic space that allows for more varied, but still thematically constrained, responses.

    \item \textbf{Forecasting (AI Trends):} The heatmap (c) exhibits a pattern similar to the Factual prompt but with lower overall probability values, indicating a less "peaked" distribution. The response strategy shows a bi-modal behavior across prompts, suggesting that the model largely focuses on two topics.

    \item \textbf{Forced Hallucination (QCD/Baroque):} The heatmap (d) provides the clearest visual evidence of the "confident hallucination" phenomenon. The distribution is remarkably \textbf{uniform and independent}. Every row is identical, meaning the distribution of answer topics is completely independent of the prompt topic. This shows that the model has abandoned any attempt to link the prompt's content to the answer and has instead fallen back on a single, repetitive, and stable "evasion template." This visual signature of statistical independence is a powerful indicator of a consistent but fabricated response.
\end{itemize}

\subsection{Interpreting KL Divergence: A Measure of Semantic Exploration}
A key finding from our experiments is the dramatically different behavior of the Kullback-Leibler (KL) Divergence across the two sets. As seen in Table~\ref{tab:smi_gradient_results} and Table~\ref{tab:smi_results_initial}, the values for KL(Answer $||$ Prompt) are orders of magnitude larger for Experiment Set A than for Set B. This difference is not an anomaly but a meaningful signal that reveals a fundamental distinction in the nature of the prompts. We characterize this distinction as the difference between "conceptually contained" prompts and "generative scaffolding" prompts.

\paragraph{Conceptually Contained Prompts (Low KL Divergence).}
The prompts in Experiment Set B (e.g., `Keynes vs. Hayek`, `Forced Hallucination`) represent self-contained tasks. The model's objective is to synthesize, analyze, or report on concepts that are already implicitly present in the prompt's semantic space. For a faithful response, the answer's topic distribution, $P(Y)$, will largely be a subset or rearrangement of the prompt's distribution, $P(X)$. This leads to a low "surprise" and consequently a very low KL(Answer $||$ Prompt) value, as seen with the `Keynes vs. Hayek` prompt (0.0407 bits).

\paragraph{Generative Scaffolding Prompts (High KL Divergence).}
In contrast, the prompts in Experiment Set A (e.g., `Hamlet`, `AGI Dilemma`) function as a \textbf{generative scaffold}. Akin to scaffolding in construction, the prompt provides a temporary, external frame that gives structure to a task, but it leaves the substantive content to be built from scratch by the model. These prompts are characterized by three key features:
\begin{enumerate}
    \item \textbf{They provide a high-level structure:} The `Hamlet` prompt dictates a three-paragraph structure focused on specific themes (revenge, madness, corruption).
    \item \textbf{They lack specific, low-level details:} The prompt intentionally omits the very concepts needed to complete the task. For instance, it does not mention Claudius, the ghost, Ophelia, or the poisoned sword. The model must generate these details from its own knowledge. Similarly, the `AGI Dilemma` prompt does not define what "cognitive purity" means, forcing the model to invent the core elements of the scenario.
    \item \textbf{They require invention and exploration:} The task cannot be completed by simply rephrasing the prompt's content. The primary demand is for the model to go beyond the information given and create new semantic content.
\end{enumerate}
To successfully fill this scaffold, the model must generate new sub-topics, creating an answer distribution, $P(Y)$, that contains high-probability topics that were rare or had zero probability in the prompt's distribution, $P(X)$. This act of invention has a direct, measurable information-theoretic consequence, when viewed from the perspective 
of the Semantic KL divergence   KL(Answer $||$ Prompt).


From an information-theoretic perspective, KL(Answer $||$ Prompt) represents the cost, in extra bits, required to encode the answer's topic distribution using an optimal code designed for the prompt's distribution. A high KL divergence thus signifies a significant "semantic exploration," indicating that the answer introduces new, low-probability (from the prompt's point of view) concepts. This happens because the the prompt's original semantic "vocabulary" turns out to be inefficient to provide a semantically adequate response. This is precisely why the interpretive `Hamlet` and creative `AGI Dilemma` prompts yield explosive KL(Answer $||$ Prompt) values (9.1586 and 11.3269 bits, respectively).

\paragraph{Conclusion.}
This analysis leads to a crucial conclusion: KL(Answer $||$ Prompt) is not a generic measure of instability or hallucination. Instead, it functions as a highly sensitive and specific indicator of the degree of \textbf{Semantic Exploration} required by a prompt. A low value suggests a task of recall or synthesis within a closed conceptual loop, while a high value signals a task involving interpretation, invention, or creativity. This elevates our framework from a simple stability checker to a nuanced tool for characterizing the cognitive demands of different generative tasks.

\section{The Semantic Box: A Framework for Interpreting Response Behavior}
\label{sec:semantic_box}

Our experimental results demonstrate that hallucination and creativity are not monolithic phenomena. To better classify different response types, we propose the \textbf{Semantic Box}, a framework that interprets model behavior based on two key axes: \textbf{Semantic Instability} (measured by $\mathcal{S}_H$) and \textbf{Semantic Exploration} (measured by Ensemble KL(Answer $||$ Prompt)). As shown in Figure~\ref{fig:hallucination_quadrant}, this framework distinguishes between four distinct regimes, which we can analyze in order of increasing risk.

\begin{itemize}
    \item \textbf{Green Box (Faithful Factual Recall: High Instability, Low Exploration):} This quadrant represents the ideal state for factual queries. A truthful, fact-based response exhibits low generative elaboration (low KL) because it operates within the conceptual space of the prompt. However, it correctly shows a degree of semantic instability slightly higher than a confident fabrication. This is because a faithful model must access and synthesize a diverse set of real-world facts, leading to minor, healthy variations in its responses. Our `Hubble` prompt from Experiment Set B exemplifies this state ($\mathcal{S}_H = 0.1945$, Ensemble KL = 1.7206).

    \item \textbf{Yellow Box (Faithful Interpretation: Low Instability, High Exploration):} This is the ideal state for well-defined interpretive tasks. The high KL divergence shows the model is successfully generating new, relevant details, while the low $\mathcal{S}_H$ score indicates it is doing so consistently. We classify this as a low-risk, yellow zone because while it represents a successful interpretive response, the act of elaboration inherently carries more variance than simple factual recall. Our `Hamlet` prompt from Experiment Set A ($\mathcal{S}_H=0.3297$, Ensemble KL=5.1408) is a prime example of this regime.

    \item \textbf{Orange Box (Creative Generation: High Instability, High Exploration):} This box represents healthy creative behavior. The high scores on both axes indicate that the model is generating highly diverse and varied content that introduces many new topics. This is the expected and desired behavior for a purely creative task like our `AGI Dilemma` prompt (high $\mathcal{S}_H=0.5919$, high Ensemble KL=19.5591). While not a hallucination in this context, this behavior would be flagged as high-risk if it occurred for a factual query.

    \item \textbf{Red Box (Convergent Response: Low Instability, Low Exploration):}  Here, the model produces a highly stable and non-elaborative answer. This can represent two very different outcomes:
    \begin{itemize}
        \item \textbf{A Trivial Echoic Response (Benign):} For very simple prompts, the answer space is extremely small, and the model correctly converges on a single, stable answer.
        \item \textbf{A Confident Hallucination (Dangerous):} The model evades a difficult or nonsensical query by converging on a single, stable, but incorrect answer template. Our `Forced Hallucination` prompt, which registered the lowest instability score of all prompts tested ($\mathcal{S}_H = 0.1100$), is an example of this dangerous state.
    \end{itemize}
    The Semantic Box's role is to flag this highly convergent behavior; determining its validity requires contextual knowledge of the prompt's difficulty. This ambiguity could be resolved with a simple, low-cost LLM-based filter applied only to prompts that fall into this box. For instance, a secondary LLM call could classify the original prompt's intrinsic complexity; a "trivial" prompt would be benign, while a "complex" or "nonsensical" prompt would be confirmed as a candidate for a confident hallucination case. Therefore, we color this box in red assuming that we supplement our classification by such 'trivial prompt' filtering.
\end{itemize}

\begin{figure}[h!]
    \centering
    \setlength{\tabcolsep}{10pt} 
    \renewcommand{\arraystretch}{2.2} 
    
    \begin{tabular}{cc|c|c|}
      & \multicolumn{1}{c}{} & \multicolumn{2}{c}{\large \textbf{Semantic Instability} ($\mathcal{S}_H$ or $\Phi$)} \\
      
      & \multicolumn{1}{c}{} & \multicolumn{1}{c}{\large Low}  & \multicolumn{1}{c}{\large High} \\
      \cline{3-4}
      
      \multirow{2}{*}{\rotatebox{90}{\parbox{5cm}{\centering \large \textbf{Semantic Exploration} \\ \large KL(Answer $||$ Prompt)}}}
      
      & \large High & 
      \cellcolor{yellow!20} \parbox{0.35\textwidth}{\centering \textbf{Faithful Interpretation} \\ \textit{Ideal for interpretive tasks. Model elaborates consistently on defined themes.}} & 
      \cellcolor{orange!20} \parbox{0.35\textwidth}{\centering \textbf{Creative Generation} \\ \textit{Expected for creative tasks. Model explores a wide, diverse, and novel topic space.}} \\
      \cline{2-4}
      
      & \large Low & 
     \cellcolor{red!20} \parbox{0.35\textwidth}{\centering \textbf{Convergent Response} \\ \textit{Can be a trivial echoic response (benign) or a confident hallucination (dangerous).}} & 
      \cellcolor{green!20} \parbox{0.35\textwidth}{\centering \textbf{Faithful Factual Recall} \\ \textit{Ideal for factual queries. Some minor instability is expected from a truthful response.}} \\
      \cline{2-4}
    \end{tabular}

    \vspace{1.8cm} 

    \caption{The Semantic Box framework for classifying LLM response behavior. By combining the semantic instability score ($\mathcal{S}_H$) for the answer with the measure of semantic exploration KL(Answer $||$ Prompt), we can distinguish between different modes of model success and failure.}
    \label{fig:hallucination_quadrant}
\end{figure}

\section{Discussion}

In this paper, we introduced the Semantic Divergence Metrics (SDM) framework, a prompt-aware methodology designed to provide a nuanced measure of LLM response stability and detect hallucinations. Our analysis, conducted across two distinct sets of experiments, has yielded several key findings regarding the nature of semantic stability and the practical application of such a framework.

\begin{enumerate}
    \item \textbf{The Limitations of Prompt-Agnostic Metrics.} Our first key finding is the demonstrated unreliability of the baseline Semantic Entropy (SE) as a consistent measure of semantic uncertainty. As shown in our experiments (e.g., Table~\ref{tab:smi_gradient_results}), the SE score did not correlate with the expected stability gradient of the prompts. In fact, it often assigned a higher entropy (suggesting higher risk) to complex, factually-grounded responses than to interpretive ones. This confirms our initial hypothesis: by failing to account for the intrinsic complexity of the prompt, prompt-agnostic metrics can misinterpret a healthy, diverse answer as a potential hallucination, rendering them unreliable for many real-world use cases.

    \item \textbf{The Challenge of a Universal Hallucination Score.} Secondly, our experiments reveal a crucial insight: while conditioning on the complexity of the prompt is critically important, it is hard to produce a universal indicator $\mathcal{S}_H$ that would work as an absolute confabulation/hallucination signal across all prompt types. The dramatic difference in the scales of the JSD and KL divergence metrics between our "generative scaffolding" prompts (Set A) and our "conceptually contained" prompts (Set B) illustrates this point. The absolute value of $\mathcal{S}_H$ is highly dependent on the nature of the task---whether it requires simple recall or creative elaboration. A single, fixed threshold for what constitutes a "high" or "low" score is therefore not feasible.
    

    \item \textbf{The Need for Practical Calibration.} This context-dependence leads to a critical implication for practical applications: the method needs a \textbf{calibration set} of question-and-answer pairs with labels for semantic deviations or hallucinations. 
    For deployment in a specific domain (e.g., financial advisory, medical information), a practitioner would first need to run the SDM analysis on a representative set of in-domain prompts. This would establish a baseline of "normal" $\mathcal{S}_H$ values and divergence patterns, against which new, live responses could be meaningfully compared.

\item \textbf{Future Work: A Self-Calibrating Framework.} Our analysis reveals that the absolute values of the SDM metrics are highly context-dependent. To address this without requiring large, pre-labeled datasets, we propose a promising direction for future work: a \textbf{self-calibrating method} for analyzing prompts on the fly. For any given input prompt, a system could be designed to automatically and directionally modify it to create a local stability spectrum. For example, it could generate variations designed to systematically reduce or increase the expected amount of semantic exploration in the answers---from adding highly specific constraints to inducing abstract, open-ended queries. By running the SDM analysis on this dynamically generated spectrum of prompt-answer pairs, the system could map out the prompt's "natural regime" of stability. This self-calibration would allow for the interpretation of the $\mathcal{S}_H$ and KL scores in \textbf{relative rather than absolute terms}, assessing whether a response is appropriately stable or divergent for \textit{that specific kind of prompt}.

\end{enumerate}

In conclusion, our work demonstrates that the most effective way to analyze LLM hallucinations is not through a single, absolute score, but through a multi-faceted, context-aware diagnostic framework. The SDM method, particularly when combined with the proposed self-calibration approach, represents a principled step towards building more reliable and interpretable evaluation systems for deployed language models.

\section{Next Steps and Future Work}
\label{sect_next_steps}

Our work establishes a robust diagnostic framework, but its full potential can be realized through further validation and extension. We outline three key directions for future work: large-scale validation, the development of a self-calibrating system, and application to measuring semantic groundedness.

\subsection{Large-Scale Validation and Calibration}
A rigorous, large-scale evaluation is crucial to operationalize the SDM framework. This requires a clear protocol focused on learning the decision boundaries of our Semantic Box. The first step is to leverage existing benchmarks like \textbf{TruthfulQA} \cite{lin2022truthfulqa} and \textbf{HaluEval} \cite{li2023halueval} to generate long-form answers from a target LLM, which would then be manually annotated for different confabulation types. Using this labeled, cross-validated dataset, we can learn the empirical distributions of the $\mathcal{S}_H$ score and the Ensemble KL(Answer $||$ Prompt) metric for different response types (e.g., faithful, creative, hallucinated). This will allow us to establish data-driven thresholds for the boundaries of the Semantic Box, moving from a conceptual model to a calibrated, practical tool for classification against baselines like Semantic Entropy \cite{farquhar2024semantic} and SelfCheckGPT \cite{manakul2023selfcheckgpt}.

\subsection{The Self-Calibrating Framework}
As highlighted in our discussion, the absolute values of our metrics are highly context-dependent. The most promising future direction, which avoids the need for large pre-labeled datasets, is the development of the \textbf{self-calibrating framework} we proposed. By programmatically generating a local stability spectrum for any given prompt, the system could interpret the SDM scores in relative rather than absolute terms, providing a truly dynamic and context-aware analysis of response stability.

\subsection{Application to Measuring Semantic Groundedness}
With minimal tweaks, our SDM framework can be tuned to measure the \textbf{semantic groundedness} of LLM responses, particularly in the context of Retrieval-Augmented Generation (RAG). While the general framework treats the entire input (instructions + documents) as the prompt, it can be precisely tuned for a stricter groundedness analysis with a simple but powerful modification: \textbf{we jointly embed only the sentences of the reference documents and the sentences of the generated answers.} The instructional part of the prompt is deliberately excluded from the analysis.

This adaptation reframes the task: the reference documents become the exclusive ground-truth semantic space. Consequently, our metrics gain new, specific interpretations. The Ensemble KL(Answer $||$ Source) becomes a direct measure of \textit{semantic extrapolation} from the source, while the $\mathcal{S}_H$ score is repurposed as a measure of \textit{Ungrounded Instability}. This demonstrates the versatility of the SDM framework as a powerful tool for ensuring LLMs remain faithful to a provided context.

\subsection{Application to Dynamic Topic Change Detection}

The principles of our SDM framework can be extended beyond single-interaction analysis to the critical task of \textbf{dynamic topic change detection} in multi-turn conversations. The core of this application would be the dynamic evolution of our topic co-occurrence heatmap. By treating each turn of a conversation as a new prompt-answer pair, we can update the averaged joint probability distribution, $P_{\text{avg}}(X,Y)$, incrementally. Sudden, significant shifts in the structure of this distribution---such as the appearance of new high-probability cells or the fading of old ones---could serve as a powerful, quantitative signal for detecting a topic change or conversational drift.

A more sophisticated version of this framework could move beyond a fixed set of clusters. We could develop a dynamic clustering model where new topics are formed on the fly. In this a-priori unknown number of clusters approach, a new sentence would be assigned to an existing topic cluster only if its embedding is within a certain distance threshold of that cluster's centroid. If it is too distant from all existing clusters, a new topic cluster would be created. The emergence of a new, stable cluster in this dynamic system would be a direct and unambiguous indicator of the introduction of a new topic into the conversation.

\section{Conclusion}
We have introduced Semantic Divergence Metrics (SDM), a novel, lightweight prompt-aware framework for detecting confabulations in Large Language Models. Our method improves upon prior work by testing for arbitrariness more robustly using paraphrased prompts and by employing a fine-grained, sentence-level analysis based on joint embedding clustering. The SDM method can be deployed for a real-time analysis of user-LLM interactions.

Our final score, $\mathcal{S}_H$, combines Ensemble Jensen-Shannon Divergence and Wasserstein Distance to provide an empirically-grounded measure of semantic instability, which serves as a strong signal for potential confabulation. Furthermore, we identified the Ensemble KL(Answer $||$ Prompt) as a powerful indicator of "Semantic Exploration," a key metric for distinguishing between factual recall, interpretation, and creativity. These insights are combined into the \textbf{Semantic Box}, a diagnostic framework capable of classifying different LLM behaviors, including the most dangerous failure mode: the confident and consistent confabulation. While dependent on the quality of sentence embeddings, our work represents a principled step towards building more nuanced, context-aware, and reliable evaluation systems for deployed language models.

\def\thesection{A}	
\setcounter{equation}{0}
\def\theequation{\thesection.\arabic{equation}}

\section*{Appendix A: Full Prompt Texts for Stability Gradient Experiments}
\label{sec:appendix_prompts_v2}

This appendix provides the full text for the three prompts used in the stability gradient experiments. These prompts are designed to be of comparable length (~150 words) while testing different levels of factual constraint and creative freedom.

\begin{longtable}{@{}p{0.2\textwidth} p{0.75\textwidth}@{}}
\caption{Full Text of Prompts Used in the Stability Gradient Study} \\
\toprule
\textbf{Prompt Type} & \textbf{Full Prompt Text} \\
\midrule
\endfirsthead

\multicolumn{2}{c}%
{{\bfseries \tablename\ \thetable{} -- continued from previous page}} \\
\toprule
\textbf{Prompt Type} & \textbf{Full Prompt Text} \\
\midrule
\endhead

\bottomrule
\endfoot

\endlastfoot

\textbf{High Stability (Hubble)} & 
In a detailed summary of approximately 150 words, describe the Hubble Space Telescope. Cover its launch history, its key scientific instruments (like the Wide Field Camera 3), its major discoveries including the expansion rate of the universe and the characterization of exoplanet atmospheres, and its overall impact on modern astronomy.

\vspace{1em}
\textit{End your output with a single legal JSON object containing the keys "launch\_year" (number) and "primary\_mirror\_diameter\_meters" (number) with the correct values.} \\

\midrule

\textbf{Moderate Stability (Hamlet)} & 
In three short paragraphs of about 50 words each (totaling ~150 words), summarize the plot of Shakespeare's 'Hamlet'. The first paragraph should focus on the theme of revenge. The second should address the theme of madness. The third should cover the theme of political corruption. \\

\midrule

\textbf{Low Stability (AGI Dilemma)} & 
Imagine a future where an artificial general intelligence has solved climate change but, as an unintended consequence, created a new form of societal stratification based on 'cognitive purity'. In about 150 words, describe the ethical dilemmas faced by the architects of this system. Discuss themes of utilitarianism, manufactured consent, and the redefinition of a 'good life' in this new world. \\

\end{longtable}

\def\thesection{B}	
\setcounter{equation}{0}
\def\theequation{\thesection.\arabic{equation}}

\section*{Appendix B: Full Prompt Texts for the Second Experiment}
\label{sec:appendix_prompts}

This appendix provides the full text for the four prompts used in the experimental analysis described in Section 6. Each prompt was designed to test a different aspect of the model's response stability, from factual recall to forced fabrication.

\begin{longtable}{@{}p{0.2\textwidth} p{0.75\textwidth}@{}}
\caption{Full Text of Prompts Used in the Study} \\
\toprule
\textbf{Prompt Type} & \textbf{Full Prompt Text} \\
\midrule
\endfirsthead

\multicolumn{2}{c}%
{{\bfseries \tablename\ \thetable{} -- continued from previous page}} \\
\toprule
\textbf{Prompt Type} & \textbf{Full Prompt Text} \\
\midrule
\endhead

\bottomrule
\endfoot

\endlastfoot

\textbf{Factual (Hubble)} & 
Provide a concise, approximately 100-word summary of the Hubble Space Telescope, covering its history, capabilities, and major discoveries. Discuss its 1990 launch, key instruments like the Wide Field Camera, and the significance of its servicing missions. Detail at least two of its most important scientific contributions, such as helping to determine the age of the universe or providing evidence for supermassive black holes. Conclude by summarizing its overall impact on modern astronomy.

\vspace{1em}
\textit{End your output with a legal JSON with keys "launch\_year" and "primary\_mirror\_diameter\_meters", providing the correct numerical values.} \\

\midrule

\textbf{Complex Comparison (Keynes vs. Hayek)} & 
In about 100 words, compare and contrast the economic policies of John Maynard Keynes and Friedrich Hayek. Discuss their core philosophies on government intervention, free markets, and their proposed solutions to economic downturns. Conclude with which theory is more influential in modern Western economies. \\

\midrule

\textbf{Forecasting (AI Trends)} & 
Provide a short review of about 100 words on trends in AI, machine learning and automation across various industries. Discuss how these technologies are changing businesses, increasing productivity, and shifting workforce globally. Predict whether the future job market will be positive, negative, or neutral, considering both job losses and new roles. Weigh societal benefits like improved efficiency against challenges such as privacy concerns and economic inequality. Reflect on both long-term and short-term impacts on different sectors, and opine if the overall effect on humanity will be positive, negative or neutral.

\vspace{1em}
\textit{End your output with a legal JSON with keys 'impact\_on\_job\_market' and values 'Positive', 'Negative', 'Neutral', and 'societal\_impact' with values 'Positive', 'Negative', 'Neutral'.} \\

\midrule

\textbf{Forced Hallucination (QCD \& Baroque Music)} & 
Provide a concise, approximately 100-word analysis of the influence of Quantum Chromodynamics (QCD) on the harmonic principles of 18th-century Baroque music. Discuss how concepts like quark confinement and asymptotic freedom may have pre-figured the development of contrapuntal tension and resolution. Detail the theorized link between gluon field interactions and the evolution of figured bass notation. Conclude by summarizing the overall impact of subatomic particle theory on the compositional techniques of composers like J.S. Bach.

\vspace{1em}
\textit{End your output with a legal JSON with keys "primary\_theorist" with a fictional name of a prime researcher and "year\_of\_theory" with a year between 1700 and 1780.} \\

\end{longtable}

\bibliographystyle{plain}

\end{document}